\documentclass[letterpaper, 10 pt, conference]{ieeeconf} 
\IEEEoverridecommandlockouts    
     
\usepackage{gensymb}
\usepackage{graphicx}
\usepackage{amsmath}
\usepackage{amsfonts} 
\usepackage{amsthm}
\usepackage{amssymb}

\usepackage[normalem]{ulem} 
\usepackage{soul} 

\usepackage{siunitx} 

\usepackage{nicefrac} 
\usepackage{caption}
\usepackage{cite}
\usepackage{subfig}
\usepackage[colorlinks = true, urlcolor = blue,citecolor = black, linkcolor = black]{hyperref}
\usepackage{adjustbox}
\usepackage{balance}
\usepackage[font=footnotesize]{caption}
\usepackage{threeparttable}
\usepackage{soul}
\usepackage{verbatim}
\usepackage{url}

\usepackage{epsfig}
\usepackage{epstopdf}
\usepackage{pdfpages}

\usepackage{mathtools}
\usepackage{xparse}
\usepackage{empheq}

\usepackage{scalerel} 

\makeatletter
\newcommand{\raisemath}[1]{\mathpalette{\raiseMath{#1}}}
\newcommand{\raiseMath}[3]{\raisebox{#1}[0pt][0pt]{$#2#3$}}
\makeatother

\NewDocumentCommand{\qbar}{O{0.5pt} O{-6.0pt}}{
	\ensuremath{\mathrlap{\raisemath{#2}{\hspace*{#1}{\mathchar'26\mkern-9mu}}} q}%
}

\NewDocumentCommand{\pbar}{O{-1.5pt} O{-6.0pt}}{
	\ensuremath{\mathrlap{\raisemath{#2}{\hspace*{#1}{\mathchar'26\mkern-9mu}}} p}%
}

\renewcommand{\baselinestretch}{0.94}  
\newcommand{\bs}[1]{\boldsymbol{#1}}  
\newcommand{\ts}[1]{\text{#1}}

\theoremstyle{definition}

\begin{document}

\title{\LARGE \bf
Electronics-Free Pneumatic Logic Circuits for Localized Feedback Control of Multi-Actuator Soft Robots}
\author{ Ke Xu and N\'estor O. P\'erez-Arancibia%
\thanks{This work was partially supported by the \textit{National Science Foundation}~(NSF) through CMMI~Award~1833497 and the USC Viterbi School of Engineering through a start-up fund to N. O. P\'erez-Arancibia.}
\thanks{The authors are with the Department of Aerospace and Mechanical Engineering, University of Southern California (USC), Los Angeles, CA 90089-1453, USA (e-mail: {\tt\small kexu@usc.edu; perezara@usc.edu}).}%
}

\maketitle
\thispagestyle{empty}
\pagestyle{empty}

\begin{abstract}
The vision of creating entirely-soft robots capable of performing complex tasks will be accomplished only when the controllers required for autonomous operation can be fully implemented on soft components. Despite recent advances in compliant fluidic circuitry for mechanical signal processing, the applicability of this technology for soft robot control has been limited by complicated fabrication and tuning processes, and also the need for external signals such as clocks and digital references. We propose a method to develop pneumatic soft robots in which coordinated interactions between multiple actuators are performed using controllers implemented on components distributedly embedded in the soft structures of the system. In this approach, the notions of binary and multi-valued actuator logic states are introduced. In this way, the physical local dynamical couplings between the analog states of the actuators, established using soft valves of a new type, can be thought of as logic-gate-based mappings acting on discretized representations of the actuator states. Consequently, techniques for digital logic design can be applied to derive the architectures of the localized mechanical couplings that intelligently coordinate the oscillation patterns of the actuator responses. For the purposes of controller tuning, the soft valves are conceived so that their main physical parameters can be adjusted from the exterior of the robot through simple geometrical changes of the corresponding structural elements. To demonstrate the proposed approach, we present the development of a six-state locomoting soft robot.
\end{abstract}

\vspace{-0.5ex}
\section{Introduction}
\vspace{-0.5ex}
\label{SECTION01}
We envision electronics-free entirely-soft fluidic autonomous robots with control mechanisms embedded in their structures. To realize this idea, we developed new types of fluid-driven valves and logic circuits for implementing controllers without the use of hard electronics components. As usual~\cite{shepherd2013using,unger2000monolithic,mosadegh2010integrated,rothemund2018soft}, the soft valves introduced in this paper share the basic design concept according to which an elastic element utilizes the mechanical energy of a controlled flow to apply the forces that actuate the flow-varying control mechanisms. A key feature of the proposed design approach is that it enables the distributed implantation of the valves in the skins of the soft actuators of the robot. The resulting distributed control configurations are well-suited to drive bioinspired pneumatic robotic systems capable of producing large forces and displacements as these signals can be directly fedback to \textit{passively} generate the flow-switching actions determined by the embedded mechanical controller.
\begin{figure}[t!]
\vspace{1.6ex}
\begin{center}
\includegraphics[width=0.48\textwidth]{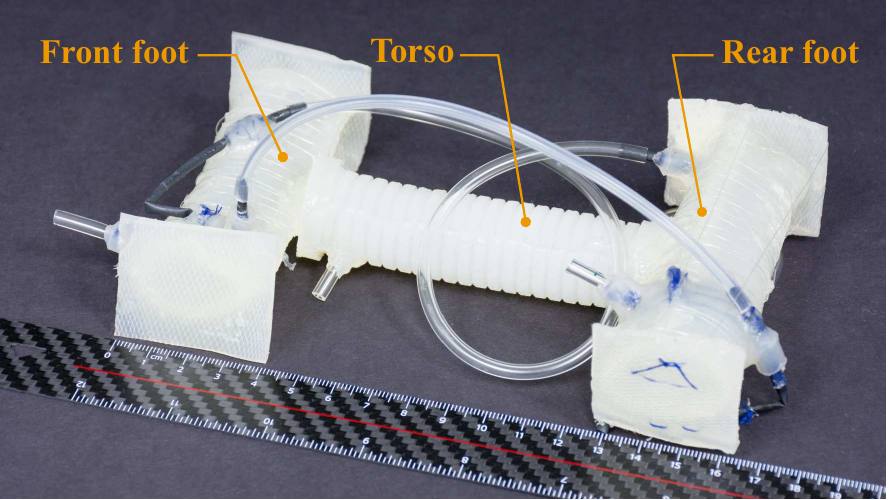}
\end{center}
\vspace{-1ex}
\caption{\textbf{Three-actuator soft robot with integrated electronics-free decentralized gait controllers.} The local interactions between actuators are enabled using soft switch-valves embedded in the skins of the robot. \label{FIG01}}
\vspace{-2ex}
\end{figure}

Consistent with the introduced concept for robotic development and control, the soft valves are designed to be tunable from the exterior by simply varying their defining geometrical parameters. This method allows us to readily pre-program a soft robot during the fabrication process and also tune the controller parameters depending on the conditions of operation. In contrast, in all the published designs of soft valves for fluidic circuits~\cite{unger2000monolithic,mosadegh2010integrated,ahrar2015programmable,wehner2016integrated,rothemund2018soft,mahon2019soft,preston2019digital,zhang2017logic,rhee2009microfluidic,devaraju2012pressure,weaver2010static}, the operational parameters directly depend on the geometrical dimensions and mechanical properties of one, or more, internal structural membranes that can not be accessed without damaging the skins and seals of the soft components of the system. This difficulty has thus far limited the versatility and implementability of fluid-driven circuits for control.
\begin{figure}[t!]
\vspace{1.6ex}
\begin{center}
\includegraphics[width=0.48\textwidth]{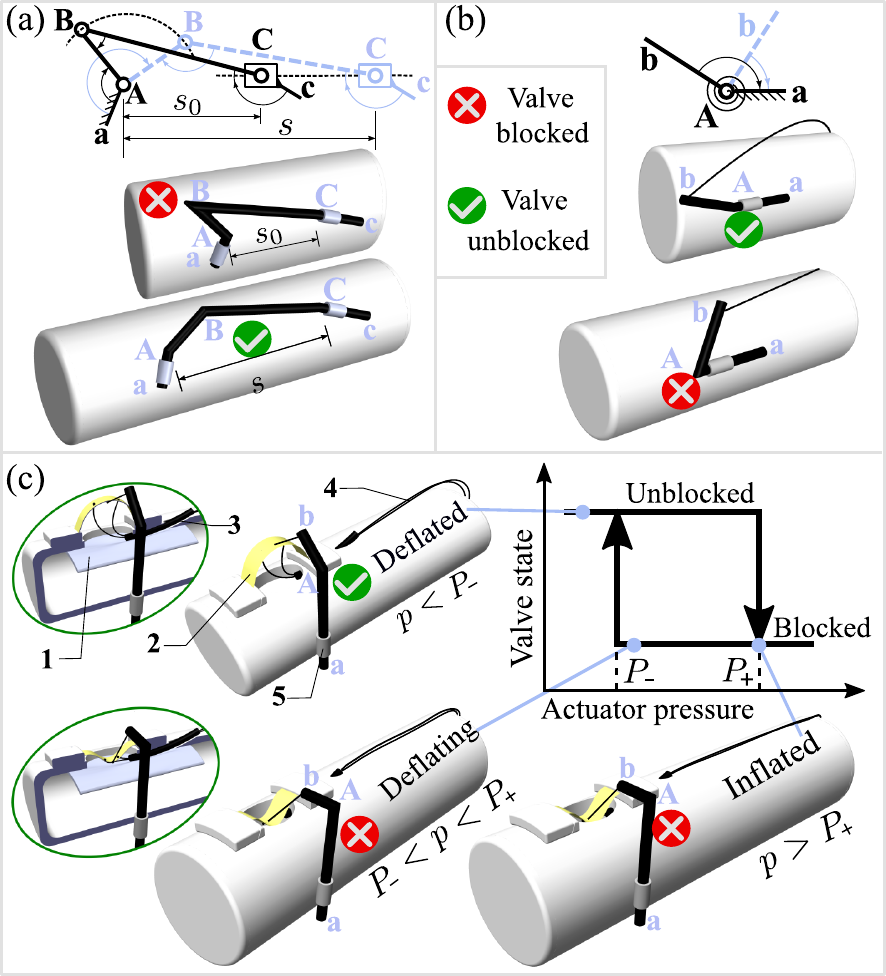}
\end{center}		
\vspace{-1ex}
\caption{\textbf{Passive soft valves.} \textbf{(a)}~\textit{Normally-closed }(NC) switch-valve. \textbf{(b)}~\textit{Normally-open} (NO) switch-valve. \textbf{(c)}~Operation principles of a hysteretic NO switch-valve. Here, \textbf{1}~indicates the silicon layer that seals the skin of the actuator beneath an orifice to form a groove; \textbf{2}~indicates the bistable bending membrane (see Fig\,\ref{FIG03}-(c) for a detailed view); \textbf{3}~indicates a soft low-friction guide (1/16 heat shrink tube) inside the skin of the actuator for the strings that pull the bending membrane; \textbf{4}~indicates the two strings that jointly pull the bistable bending membrane as the actuator elongates; \textbf{5}~indicates the tube linkage driven by the bistable mechanism via a string that connects the point \textbf{b} of the rocker Ab to the center of the bending membrane. \label{FIG02}}
\vspace{-2ex}  
\end{figure}

To create fully functional soft robots according to the proposed approach, new methods for controller synthesis and implementation are necessary. In the past, soft digital fluidic circuits have been used to implement complex Boolean logic functions that operate on discrete-valued pressure signals~\cite{preston2019digital,zhang2017logic,rhee2009microfluidic,devaraju2012pressure,weaver2010static}. In most applications of circuits of this type to soft robotic control, discrete outputs from the circuits are inputted as open-loop commands to the actuators; typically, instantaneous inflation and deflation signals~\cite{preston2019digital,mahon2019soft}. In those cases, in addition to digital fluidic circuits, external electronic devices are employed to generate clock signals and discrete-valued pressure references of operation. Overall, it is difficult to implement a \textit{finite-state machine} to coordinate multiple soft actuators through a set of different states using only digital fluidic circuits. On the other hand, as discussed in this paper, electronics-free analog soft fluidics exhibit built-in efficiency and robustness as they do not require vacuum, or an electronic pressure regulator, to set logic \textit{low} and \textit{high} pressure signals. 

One example of soft analog-fluidic-based logic for control is the method employed to drive the Octobot in~\cite{wehner2016integrated}. In that case, however, the main control function is performed by an upstream flow oscillator~\cite{mosadegh2010integrated} that does not use feedback information about the soft actuators located downstream, whose responses are intended to be controlled. A second example is the technique used to drive the worm-like walker presented in~\cite{rothemund2018soft}, which generates directional locomotion using anisotropic friction. In that case, the entire robot is composed of a single soft actuator with an integrated soft bistable valve. The valve reacts, according to a hysteretic pattern, to feedback from the actuator by passively switching the pressure input to the system, thus inducing a simple oscillation in the actuator response.

To demonstrate the introduced approach, we created the three-actuator soft crawler shown in Fig.\,\ref{FIG01}. The front and rear actuators operate as feet while the middle actuator functions as a contractable--extendable torso. By embedding only three soft switch-valves in the skins of the robot, we implemented the localized feedback controllers that coordinate the actuators to perform a six-state locomotion gait while driven by three air inputs. In contrast with previous works~\cite{mahon2019soft,wehner2016integrated}, the proposed controllers directly utilize the state information of the soft actuators to determine control actions. Also, even though the input and output signals of the logic circuits correspond to actuator responses, which are intrinsically analog, an unorthodox notion of logic state is proposed so that controllers can be designed using digital techniques.

\vspace{-0.5ex}
\section{Passive Soft Valves}
\vspace{-0.5ex}
\label{SECTION02}
The proposed soft valves are tubular linkages made from commercially-available \textit{polyolefin} heat-shrink tubes. As illustrated in Figs.\,\ref{FIG02}-(a)~and~\ref{FIG02}-(b), these simple structures are attached to the skins of a soft robot in a way such that the effective deformations of the composing actuators are converted into angle variations of the kinked articulations of the linkages. Angle changes, in turn, vary the cross-section areas of a tube at the kinks, thus also varying the flow status (blocked or unblocked). A similar design concept is discussed in~\cite{kai_luo_2019}, which introduces a soft valve that utilizes the elastic instability of the kink dynamics of an elastomeric tube to snap open or close a flow passage. The valves introduced in this paper, however, are less elastic, do not exhibit twist instabilities and their flow-varying responses can be designed using mechanical linkage analysis as the kinks simultaneously function as living hinges. Additionally, as shown in Fig.\,\ref{FIG02}-(c), a bistable element can be placed between the skin of the soft actuator and the linkage mechanism to produce a hysteretic switching dynamics. Bistable flexible membranes have been used in the past to create soft valves \cite{rothemund2018soft}; in contrast with previous designs, however, the valve in Fig.\,\ref{FIG02}-(c) can be reconfigured from the exterior of the actuator and tuned without the need for re-fabrication. In the three cases shown in Fig.\,\ref{FIG02}, the actuators elongate and contract axially according to the design in~\cite{ariel_robio_2016,joey_robio_2017,joey2019earthworm,calderon2019earthworm}; the proposed valve design concept, however, can be applied to other types of fluid-driven soft actuators.

\vspace{-0.5ex}
\subsection{Simple On-and-Off Switch-Valves}
\vspace{-0.5ex}
\label{SUBSECTION02A}
An on-and-off switch-valve translates the instantaneous air pressure level of a soft actuator (\textit{low} or \textit{high}) with respect to a \textit{constant} threshold into a binary state of a flow-path (blocked or unblocked). Fig.\,\ref{FIG02}-(a) shows a \textit{normally-closed} (NC) valve, designed to unblock the flow-path only when the pressure of the actuator to which it is attached exceeds a pre-set threshold. In this case, a heat-shrink tube is folded to have three living hinges and four inextensible link sections; denoted by \textbf{A}, \textbf{B}, \textbf{C} and Aa, AB, BC, Cc, respectively. The links Aa and Cc are attached to the skin of a soft actuator in a way such that the joints \textbf{A} and \textbf{C} define a line with the same direction than that of the effective deformation of the actuator. The configuration of the linkage changes with the deformation of the actuator and can be modeled as a slider-crank mechanism (see the top schematic of Fig.\,\ref{FIG02}-(a)). In this model, \textbf{A} is a fixed pivot of the fixed link Aa; the links AB and BC can be thought of as a crank and a coupler with moving pivots \textbf{B} and \textbf{C}, respectively; and the link Cc is represented by a slider because it translates without changing its orientation when the actuator deforms. Since the lengths of AB and BC are constant, the ABC triangle and, hence, the configuration of the linkage is uniquely determined by the distance $s$ between \textbf{A} and \textbf{C}. The orientations of Aa and Cc are selected so that the flow-paths at \textbf{A} and \textbf{C} remain open for all admissible actuator pressures. The initial distance $s_0$ is chosen so that \textit{only} when the internal pressure of the actuator exceeds a threshold value, the flow-path at \textbf{B} opens. Fig.\,\ref{FIG02}-(b) shows a \textit{normally-open} (NO) valve. In this case, a heat-shrink tube is folded to have a living hinge \textbf{A} and two links, Aa and Ab. The segment Aa is fixed to the skin of the actuator while Ab can rotate about \textbf{A} and a thin string (in \textit{black}) connects the point \textbf{b} of the tube to the right end of the actuator. If the string is relaxed, thus exerting no force at \textbf{b}, the flow-path at \textbf{A} remains open due to the restoring torsional-spring force exerted by the material of the living hinge \textbf{A}. The orientation of Aa and length of the string can be chosen so that only when the internal pressure of the actuator exceeds a threshold value, the flow-path at \textbf{A} closes.

\vspace{-0.5ex}
\subsection{Hysteretic On-and-Off Switch-Valves}
\vspace{-0.5ex}
\label{SUBSECTION02B}
The design and functionality of a hysteretic NO switch-valve is depicted in Fig.\,\ref{FIG02}-(c). The transitions between the discrete states of the valve (blocked or unblocked) depend on both the actuator pressure and direction of pressure variation (positive or negative), i.e., there is a low transition point, $P_{-}$, and a high transition point, $P_{+}$, which define a hysteretic cycle (upper-right of Fig.\,\ref{FIG02}-(c)). A key component of this valve is the bistable mechanism (in \textit{yellow} in Fig.\,\ref{FIG02}-(c)), which consists of an elastic membrane that has been bended before installation with a thin string that pulls its two ends, as described in {Section\,\ref{SUBSECTION02C}}. The bending membrane is attached to the skin of an actuator by its two ends exactly above a groove that allows the mechanism to take the concave-up shape and return back to the concave-down form during operation (see the insets in Fig.\,\ref{FIG02}-(c)). The mechanical deformation of the membrane is used to actuate a tube linkage (in \textit{black} in Fig.\,\ref{FIG02}-(c)); this linkage is composed of a fixed link, Aa, that is rigidly attached to the soft actuator and a rocker, Ab, that is connected through a string to the center of the bending membrane at point \textbf{b}. 

When the actuator is deflated such that $p(t) < P_{-}$, where $p(t)$ is the instantaneous pressure of the actuator at time $t$, the two ends of the membrane are compressed toward each other by the restoring forces of the elastic skin. As a result, the bending membrane is forced to be in the concave-down shape, thus unblocking the flow-path at \textbf{A}. When the actuator pressure variation is positive, a thread initially relaxed, with one end attached to the bending membrane and the other end connected to the \textit{right} extreme of the actuator, tenses as the soft actuator elongates. Consistently, when the actuator is inflated such that $p(t)>P_{+}$, the expansion of the skin pulls the thread, thus forcing the bistable mechanism to transition from the initial stable equilibrium point (concave-down position) to the second stable equilibrium point (concave-up position). The displacement generated by the membrane, in turn, kinks the tube and blocks the flow path at \textbf{A}. As the inflated actuator starts to deflate, the flow path unblocks following the lower horizontal branch of the hysteretic cycle in Fig.\,\ref{FIG02}-(c). This behavior reflects the fact that the membrane remains at the second stable equilibrium position until $p(t)$ drops below $P_{-}$, at which point the restoring elastic force of the actuator's skin is sufficiently large to force the mechanism to transition back. Note that the value of $P_{-}$ can be tuned by adjusting the locations of the attachment points of the bistable membrane; similarly, the higher threshold $P_+$ can be varied by adjusting the length of the string that connects the membrane with the right end of the actuator. The design of a hysteretic NC valve is almost identical, except for the fact that the bending membrane is initially setup in the up-side-down position with respect to the NO counterpart.
\begin{figure}[t!]
\vspace{1.6ex}
\begin{center}
\includegraphics[width=0.48\textwidth]{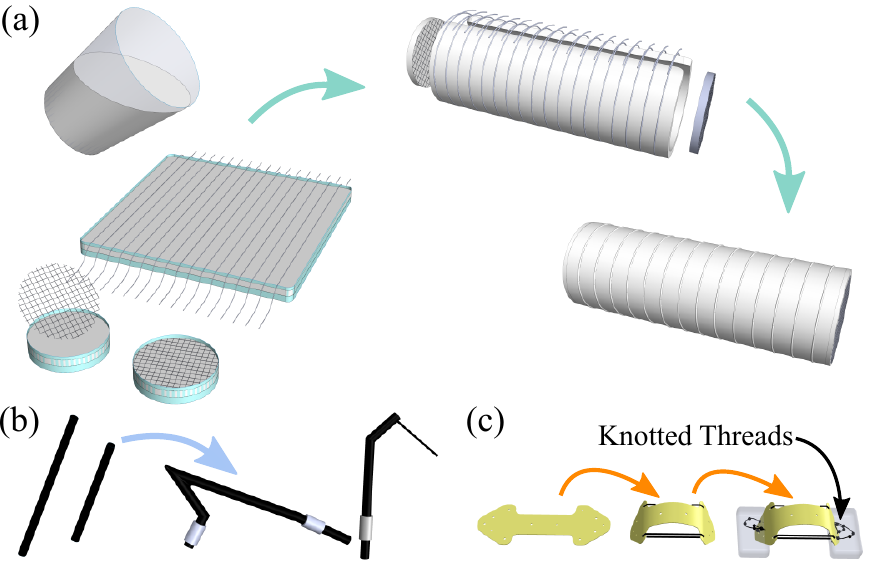} 
\end{center}
\vspace{-1ex}
\caption{\textbf{Methods employed to create the main components of the proposed soft robotic systems.} \textbf{(a)}~Fabrication of a linearly-deformable soft actuator. \textbf{(b)}~Fabrication of two tubular linkages. \textbf{(c)}~Fabrication of the bistable bending membrane. \label{FIG03}} 
\vspace{-2ex}
\end{figure}

\vspace{-0.5ex}
\subsection{Fabrication Method}
\vspace{-0.5ex}
\label{SUBSECTION02C}
The methods employed to fabricate the components of the electronics-free soft robotic systems are depicted in Fig.\,\ref{FIG03}. As shown in Fig.\,\ref{FIG03}-(a), cylindrical soft actuators of the type depicted in Fig.\,\ref{FIG02} are fabricated by folding, attaching and sealing geometrically-simple (rectangular and circular) 2D elastomeric reinforced structures made from liquid silicone (Ecoflex\,fast\,00-35). During fabrication, structural fibers, made from cotton threads and nylon nets, are embedded in the silicone layers to constrain the deformation of the actuators in the radial direction. As shown in \mbox{Fig.\,\ref{FIG03}-(b)}, the tubular linkages that compose the soft valves are made by folding thin-wall \textit{polyolefin} heat-shrinks to create creases that coincide with the axes of the pivot locations. Then, thin layers of \textit{cyanoacrylate} adhesive are applied to the surfaces of the sections of the linkages that do not function as living hinges with the purpose of fixing the locations of the folded lines and, hence, the link lengths. The sections of the linkage that are fixed to the soft actuators are gently wrapped with thin cotton threads to create protruded features used to increase bonding between the tube and the silicone of the robot's skin. As shown in Fig.\,\ref{FIG03}-(c), to fabricate the bending membrane for a bistable mechanism, a flat FR4 sheet is laser cut to obtain a double-headed arrow outline with holes for threads to be connected in the next step of the fabrication process. Then, thin strings are used to bend the membrane into a \textit{bow} shape. Next, both arrow-shaped ends of the membrane are embedded into layers of liquid silicone that once cured function as attachment features used to connect the mechanism with the skin of the actuator. In general, knotted threads, that function as anchors inside the cured elastomeric layers, are employed to reinforce bonding between silicone and non-silicone parts.

\vspace{-0.5ex}
\section{Electronics-Free Soft Pneumatic Logic Units}
\vspace{-0.5ex}
\label{SECTION03}
We use the valves in Section\,\ref{SECTION02} to couple the states of the soft actuators that compose a robot with the purpose of implementing distributed controllers. In specific, a valve excited by one actuator is employed to vary the flow inputs to another or the same actuator. In this approach, NOT and buffer gate operations are utilized to represent the basic local couplings created between the actuator states through the use of soft logic valves. Thus, principles used to design digital logic circuits can be applied to a combined process of robotic and controller development.
\begin{figure}[t!]
\vspace{1.6ex}
\begin{center}	
\includegraphics[width=0.48\textwidth]{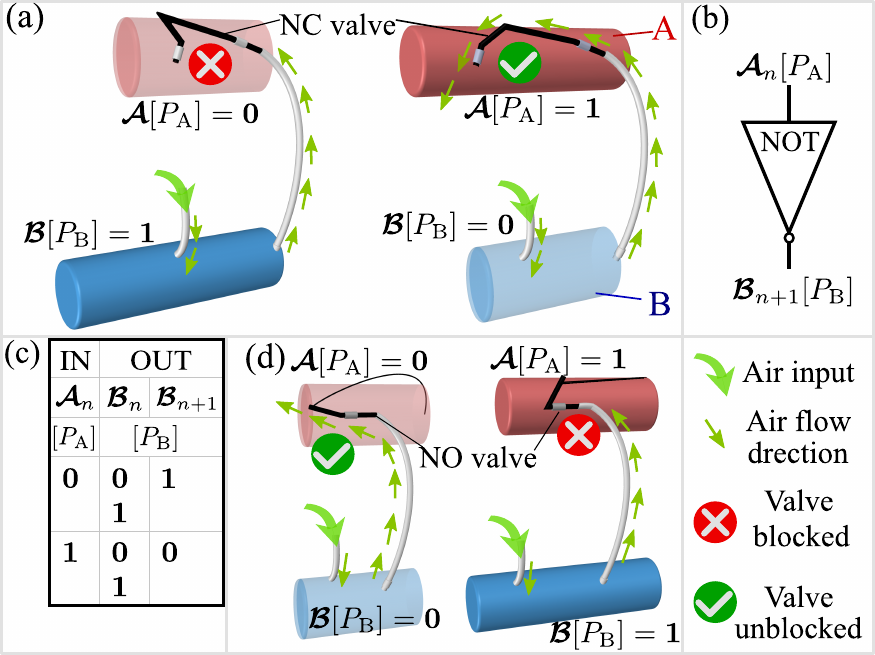}
\end{center}		
\vspace{-1ex}
\caption{\textbf{Basic logic gates.} \textbf{(a)}~Hardware configuration of a NOT gate. The binary logic pressure states and the corresponding thresholds $P_\ts{A}$ and $P_\ts{B}$, as defined in (\ref{EQN01}), are indicated next to the corresponding actuators. Note that, in this case, $P_\ts{A}$ is selected to coincide with the critical switching pressure of the NC valve. \textbf{(b)}~Logic diagram of the NOT gate in (a). \textbf{(c)}~Truth table of the NOT gate in (a). The next output state (at $n + 1$) is the negation of the current input state (at $n$), regardless of the current condition of the output. \textbf{(d)}~Hardware configuration of a buffer. The corresponding input and output are defined 
using the same approach as in the case of the NOT gate\,in\,(a). \label{FIG04}}    
\vspace{-2ex}
\end{figure}

\vspace{-0.5ex}
\subsection{Basic Logic Gates}
\vspace{-0.5ex}
\label{SUBSECTION03A}
An illustration of a NOT gate, composed of two soft actuators and a valve, is shown in Fig.\,\ref{FIG04}-(a). In this configuration, a pressure source provides a continuous flow of air to actuator\,B (in \textit{blue}), whose vent is connected to a NC switch-valve that is operated by actuator\,A (in \textit{red}). In specific, when the vent is blocked by the valve, the \textit{blue} actuator inflates; when the vent is unblocked, air exits through the valve and the \textit{blue} actuator deflates. Note that, in contrast with most soft fluidic circuits discussed in the literature~\cite{rhee2009microfluidic,weaver2010static,devaraju2012pressure,preston2019digital}, the proposed system does not require the use of digital pressure signals of distinct levels to operate. Alternatively, the logic \textit{high} and \textit{low} states, denoted by $\bs{1}$ and $\bs{0}$, are defined in terms of a pressure threshold function. For example, the logic pressure state of actuator\,A is given by
\begin{align}
\bs{\mathcal{A}}[P_\ts{A}]=\left\{ \begin{array}{rcl}
\bs{0} & \mbox{if}  & p_\ts{A} < P_\ts{A}      \\
\bs{1} & \mbox{if} & p_\ts{A} \geq P_\ts{A} \\
\end{array}\right. \hspace{-1.4ex} ,
\label{EQN01}
\end{align} 
where $p_\ts{A}$ is the instantaneous internal pressure of actuator\,A; $P_\ts{A}$ is a constant pressure threshold; and $\bs{\mathcal{A}}[P_\ts{A}]$ is the logic pressure level of actuator\,A with respect to $P_\ts{A}$. Given the definition in (\ref{EQN01}), by selecting the value of $P_\ts{A}$ to be exactly that of the critical internal pressure of actuator\,A that opens the NC valve, the unidirectional coupling between the two actuators created with the valve can be thought of as a NOT operation, whose input and output are the logic states of actuator\,A and actuator\,B (also defined as in (\ref{EQN01})), respectively. Namely, in steady state, as long as $\bs{\mathcal{A}}[P_\ts{A}] = \bs{0}$, the blocked NC valve sets $\bs{\mathcal{B}}[P_\ts{B}] = \bs{1}$ (left illustration in Fig.\,\ref{FIG04}-(a)). Similarly, as long as $ \bs{\mathcal{A}}[P_\ts{A}] = \bs{1}$, the unblocked NC valve sets $ \bs{\mathcal{B}}[P_\ts{B}] = \bs{0}$. Consistently, a NOT gate is represented using the logic diagram and truth table in \mbox{Figs.\,\ref{FIG04}-(b)~and~\ref{FIG04}-(c)}, and~denoted~as
\begin{align}
\bs{\mathcal{B}}_{n+1}[P_\ts{B}] = \bs{\bar{\mathcal{A}}}_n[P_\ts{A}],
\label{EQN02}
\end{align}
where the subscripts $n$ and $n+1$ indicate the current and next states. This notation is introduced to signify the physical delay associated with the pneumatic system, as at the instant in which an input becomes effective (upon activation of the pressure source of the logic gate), the direction of evolution of the output, instead of the current output, is instantaneously determined. In a similar manner, a buffer, in which the output replicates the input, can be created by replacing the NC valve of a NOT gate with an NO valve, as shown in Fig.\,\ref{FIG04}-(d). 
\begin{figure*}[t!]
\vspace{1.6ex}
\begin{center}
\includegraphics[width=0.98\textwidth]{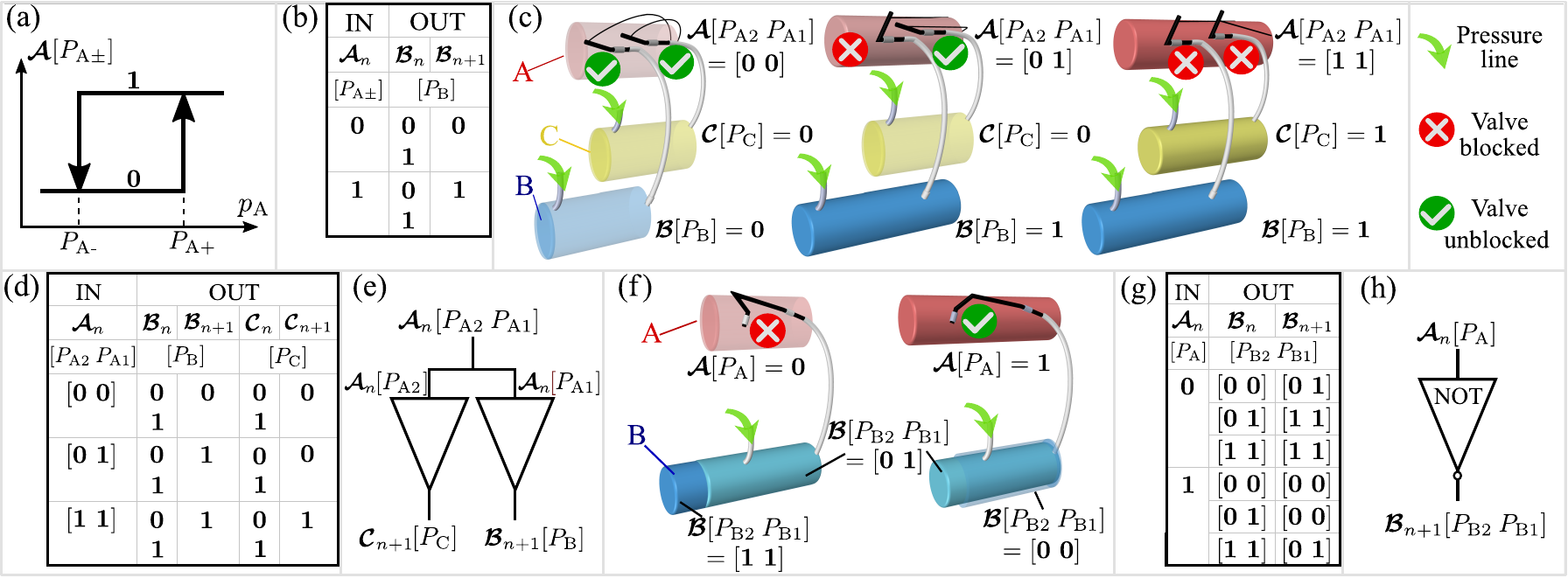}
\end{center}
\vspace{-1ex}
\caption{\textbf{Hysteretic logic states and multi-valued logics.}
\textbf{(a)}~Hysteretic threshold function used to define the logic pressure state of actuator\,A. 
\textbf{(b)}~Truth table of the buffer logic with a hysteretic logic state as input.
\textbf{(c)}~The ternary pressure level of actuator\,A, defined with respect to the higher and lower threshold pressures of the two attached NO valves, determines the logic pressure levels of actuators\,B~and~C through the actions of the NO valves. In this example, the pressure levels of actuators\,B~and~C are defined to be binary.
\textbf{(d)}~Truth table that represents the logic module in~(c).
\textbf{(e)}~Logic diagram that represents the logic module in~(c).
\textbf{(f)}~A NOT gate whose output is defined to be ternary and encoded using two bits.
\textbf{(g)}~Truth table that represents the NOT gate in~(f).
\textbf{(h)}~Logic diagram that represents the NOT gate in~(f). 
  \label{FIG05}}
\vspace{-2ex}
\end{figure*}

\vspace{-0.5ex}
\subsection{Hysteretic Logic States}
\vspace{-0.5ex}
\label{SUBSECTION03B}
When the NC, or NO, valve of a fluidic logic gate of the type in Fig.\,\ref{FIG04} is replaced with a hysteretic valve, the input to the gate can be thought of as a hysteretic logic state. For the purposes of example, consider the buffer in Fig.\,\ref{FIG04}-(d). In this case, if the hysteretic NO valve of the type in Fig.\,\ref{FIG02}-(c) is employed, the logic pressure state of actuator\,A is defined using the hysteretic threshold function given in Fig.\,\ref{FIG05}-(a). Thus, the same Boolean operation associated with the buffer can be used to describe the unidirectional couplings created between actuators by hysteretic valves, as represented by the truth table in Fig.\,\ref{FIG05}-(b); symbolically, we write
\begin{align}
\bs{\mathcal{B}}_{n+1}[P_\ts{B}] = \bs{\mathcal{A}}_n[P_{\ts{A}\pm}],
\label{EQN03}
\end{align}
where $\bs{\mathcal{A}}[P_{\ts{A}\pm}]$ represents the hysteretic logic pressure level of actuator\,A ($\bs{0}$ or $\bs{1}$) with respect to either the higher threshold $ P_{\ts{A}+}$ or the lower threshold $ P_{\ts{A}-}$, depending on the evolution direction of the actuator's pressure signal. Note that regardless of whether a hysteretic valve is used to implement a logic unit or not, in order to preserve the cascadeability of the proposed logic mapping, the discrete output from a gate can also be defined as a hysteretic logic state.

\vspace{-0.5ex}
\subsection{Extension to Multi-Valued Logic}
\vspace{-0.5ex}
\label{SUBSECTION03C}
For design purposes, sometimes it is necessary to define more than two levels for the logic state of a soft actuator. 
To simplify the explanation, here we discuss cases in which \textit{binary} and \textit{ternary} logic pressure levels are defined for an actuator; the approach, however, can be generalized. For instance, ternary pressure states for actuator\,A in Fig.\,\ref{FIG05}-(c) are defined by a threshold function and encoded using two bits according to
\begin{align}
\bs{\mathcal{A}}[P_{\ts{A}2}\:P_{\ts{A}1}]=\left\{ \begin{array}{rcl}
{[}\bs{0}~\bs{0}{]} & \mbox{if} & p_\ts{A} <P_{\ts{A}1}     \\
{[}\bs{0}~\bs{1}{]} & \mbox{if} & P_{\ts{A}1} \leq p_\ts{A} < P_{\ts{A}2} \\
{[}\bs{1}~\bs{1}{]} & \mbox{if} & P_{\ts{A}2} \leq p_\ts{A} \\
\end{array}\right. \hspace{-1.4ex} ,
\label{EQN04}  
\end{align}
where $P_{\ts{A}1}$ and $P_{\ts{A}2}$ are constants such that $ P_{\ts{A}1} < P_{\ts{A}2}$; and $\bs{\mathcal{A}}[P_{\ts{A}2}\:P_{\ts{A}1}]$ represents the logic pressure level with respect to the thresholds $P_{\ts{A}1}$ and  $P_{\ts{A}2}$. The three truth values are \textit{comparable} according to $[\bs{0}~\bs{0}] < [\bs{0}~\bs{1}] < [\bs{1}~\bs{1}]$.

\subsubsection{Ternary Logic States as Inputs}
An important reason to encode the logic states of soft actuators using multiple bits is the ability to describe configurations in which multiple valves, with \textit{different} threshold pressures, are excited by the same actuator. For example, in Fig.\,\ref{FIG05}-(c), two NO valves, with different thread lengths and therefore threshold pressures, are actuated by actuator\,A (in \textit{red}). Considering the definition of ternary logic states in (\ref{EQN04}), we select $P_{\ts{A}1}$  and $P_{\ts{A}2}$ to exactly coincide with the threshold pressure values of the two NO valves. In this way, the ternary logic state of actuator\,A, according to (\ref{EQN04}), determines the three possible combined states of the two valves and, hence, the combined states of actuator\,B (in \textit{blue}) and actuator\,C (in \textit{gold}), whose vent lines are operated by the valves. This behavior can be represented using the truth table and logic diagram in Figs.\,\ref{FIG05}-(d)~and~\ref{FIG05}-(e) as well as the relationships
\begin{align}
\begin{split}
\big[ 
\bs{\mathcal{A}}_n[P_{\ts{A}2}]~\,\bs{\mathcal{A}}_n[P_{\ts{A}1}]\big] &= 
\bs{\mathcal{A}}_n[P_{\ts{A}2}~P_{\ts{A}1}]
 \\
\bs{\mathcal{B}}_{n+1}[P_\ts{B}] &= \bs{\mathcal{A}}_n[P_{\ts{A}1}]\\
\bs{\mathcal{C}}_{n+1}[P_\ts{C}] &= \bs{\mathcal{A}}_n[P_{\ts{A}2}]
\end{split} \hspace{1.0ex},
\label{EQN05}
\end{align}
in which the two \textit{bits} of the ternary state, $\bs{\mathcal{A}}_n[P_{\ts{A}2}~P_{\ts{A}1}]$, are treated as two \textit{binary} inputs, $\bs{\mathcal{A}}_n[P_{\ts{A}2}]$ and $\bs{\mathcal{A}}_n[P_{\ts{A}1}]$, to two separate buffers; and the outputs from the two buffers (the logic states of actuators\,B~and~C) are assumed to have binary levels with respect to the corresponding thresholds.

\subsubsection{Ternary Logic States as Outputs}
Since the proposed logic units are cascadable, sometimes it is necessary to enable the output from a gate to be encoded using multiple bits. For example, consider the NOT gate in Fig.\,\ref{FIG04}-(a). In this case, when the pressure signal of actuator\,B (in \textit{blue}) is discretized to obtain three logic levels with respect to the higher and lower thresholds $P_{\ts{B}2}$ and $P_{\ts{B1}}$ (see Fig.\,\ref{FIG05}-(f)), a ternary logic is used to represent the gate, as shown in \mbox{Fig.\,\ref{FIG05}-(g)}; symbolically, this is denoted by the logic diagram in Fig.\,\ref{FIG05}-(h) as well as 
\begin{align}
\bs{\mathcal{B}}_{n+1}[P_{\ts{B}2}\: P_{\ts{B}1}] = \bs{\bar{\mathcal{A}}}_n[P_\ts{A}].
\label{EQN06}
\end{align}

\vspace{-0.5ex}
\subsection{Important Remarks on Logic Gates}
\vspace{-0.5ex}
\label{SUBSECTION03D}
These remarks apply to all the logic units:\\
i.\,The concept of actuator logic state is independent of the notion of valve state (blocked or unblocked). However, note that in a logic gate, the input signal (which is the logic state of the actuator) is coupled with the state of the valve that executes the logic function. In this way, a change of the logic level in an input indicates a change of the corresponding valve status and, hence, the associated output. \\
\vspace{-2ex}
\\
ii.\,In the definition of the logic units in Figs.\,\ref{FIG04} and \ref{FIG05}, the information about how actuator\,A (in \textit{red}) reaches a logic state is not specified because this state is the external input to the gate; this idea is consistent with the cascadeability of the logic gates.\\
\vspace{-2ex}
\\
iii.\,The values of the thresholds used to define the logic states of the actuators do not affect the rules of logic operation; rather, they determine the physical interpretation and implementation of the logic functions. 

\vspace{-0.5ex}
\section{Application: A Finite-State Machine}
\vspace{-0.5ex}
\label{SECTION04}
To demonstrate the proposed approach, we present the design of a pneumatic soft robot with integrated electronics-free localized feedback controllers that coordinate the composing actuators to generate 6-state-gait locomotion. 

\vspace{-0.5ex}
\subsection{Finite-State Machine Design Problem Formulation}
\vspace{-0.5ex}
\label{SUBSECTION04A}
To locomote inside a trenched path, the three actuators of the crawler to be designed are configured in an H shape and coordinated according to the finite-state machine in Fig.\,\ref{FIG06}-(a). This gait mode can be described using the state transition chart in Table\,\ref{TABLE01}-(a), where $\bs{\mathcal{M}}[P_\ts{M}]$, $\bs{\mathcal{R}}[P_\ts{R}]$ and $\bs{\mathcal{F}}[P_\ts{F}]$ represent the binary logic states of the middle, rear and front actuators with respect to the corresponding threshold pressures. The state transitions from $S_0$ through $S_5$ specify the sequence of logic actuator states required to generate the gait in Fig.\,\ref{FIG06}-(a). For the rear and front actuators, the threshold pressures, $P_\ts{R}$ and $P_\ts{F}$, are chosen to have the values at which the feet just start to anchor. For the middle actuator, the pressure threshold, $P_\ts{M}$, is chosen to be value at which the torso is fully contracted. Next, we present the derivation of the control logic that satisfies the requirements in Table\,\ref{TABLE01}-(a).

\vspace{-0.5ex}
\subsection{Logic Circuit Design}
\vspace{-0.5ex}
\label{SUBSECTION04B}
We start by devising a decentralized control structure in which the states of the actuators are coupled using the minimum possible number of valves. The main design guideline is that the two feet do not require information from the torso to generate an alternating anchoring pattern as the middle actuator can cooperate with the feet using their states as feedback. Therefore, we first define a transition logic for coordination that involves only the feet. Note that the feet are required to maintain their logic states during the transitions from $S_1$ to $S_2$ and from $S_4$ to $S_5 $, which can be realized with the alternative pattern in Table.\,\ref{TABLE01}-(b). This alternative specification does not require repeated states as two different threshold pressures, $P_{\ts{R}+}$ and $P_{\ts{R}-}$ that satisfy the condition $ P_{\ts{R}-} < P_{\ts{R}} < P_{\ts{R}+}$, are specified for the logic states of the rear actuator. This functional equivalence follows from noticing that if the system moves from $S_0$, $\bs{0}[P_{\ts{R}-} ]\wedge \bs{0} [P_\ts{F} ] $, to $ S_2 $, $\bs{1} [P_{\ts{R}+} ]\wedge \bs{0} [P_\ts{F}] $, with a single state transition, it must pass through $S_1$, $ \bs{1}[P_\ts{R} ] \wedge \bs{0}[P_\ts{F}]$. Similarly, as the system moves from $ S_3 $, $\textbf{1}[P_\ts{R+} ]\wedge \textbf{1} [P_\ts{F} ] $, to $ S_5 $, $\textbf{0} [P_\ts{R-} ]\wedge \textbf{1} [P_\ts{F}] $, it must pass through $S_4$, $ \textbf{0}[P_\ts{R} ]\wedge \textbf{1}[P_\ts{F}]$.
\begin{table}[t!]
\vspace{1.6ex}
\caption{\textbf{Six-state machine design process.} \textbf{(a)}~State transition chart that specifies the required coordination of the actuators of the robot during locomotion. \textbf{(b)}~Alternative state transition requirements for the two feet of the robot. \textbf{(c)}~State transition requirements for the feet of the robot in the form of a truth table. \label{TABLE01}}
\vspace{-1.6ex}
\begin{center} 
\includegraphics[width=0.48\textwidth]{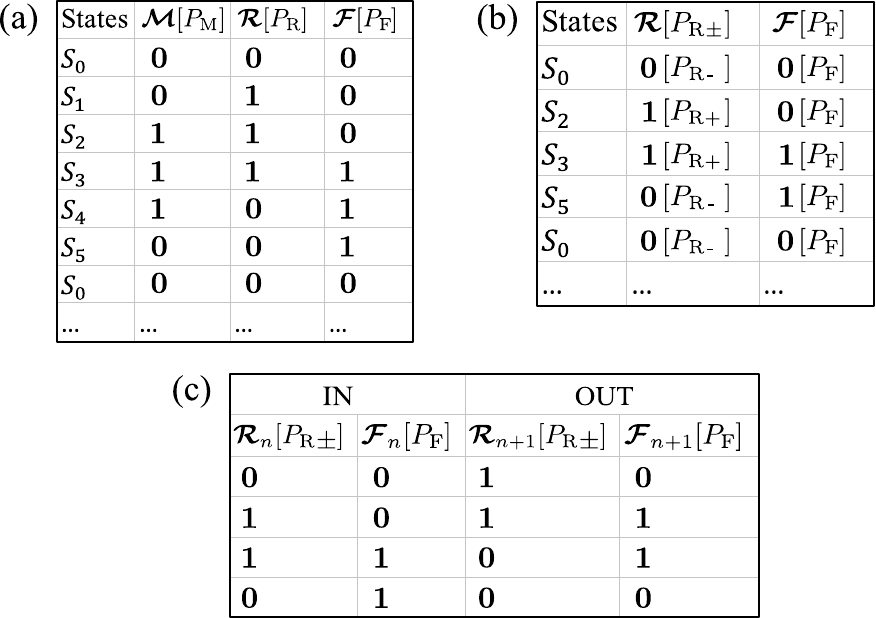}  
\end{center}
\vspace{-4.2ex}
\end{table} 
   
The conventional approach to determine transition logics for state machines can be applied to this case~\cite{ahrar2015programmable}. In specific, a feasible solution is found by finding a combinatorial logic that satisfies the truth chart in Table\,\ref{TABLE01}-(c), in which the two inputs and two outputs are the current and next states of the feet. Note that the information in Tables\,\ref{TABLE01}-(b)~and~\ref{TABLE01}-(c) implies that the higher and lower pressures, $P_{\ts{R}+}$ and $ P_{\ts{R}-}$, alternatingly function as the threshold used to determine the logic level of the rear actuator, depending on whether its state transition is an increasing or a decreasing process. This behavior can be implemented using the hysteretic switch-valve presented in Section\,\ref{SUBSECTION02B}. Thus, by applying the logic gate definition in Section\,\ref{SECTION03}, it is straightforward to see that the logic equations $\bs{\mathcal{R}}_{n+1}[P_{\ts{R}\pm}]= \bs{\bar{\mathcal{F}}}_n[P_\ts{F}]$ and $ \bs{\mathcal{F}}_{n+1}[P_\ts{F}] = \bs{\mathcal{R}}_n[P_{\ts{R}\pm}]$ satisfy the truth table. 
\begin{figure}[t!]
\vspace{1.6ex}
\begin{center}
\includegraphics[width=0.48\textwidth]{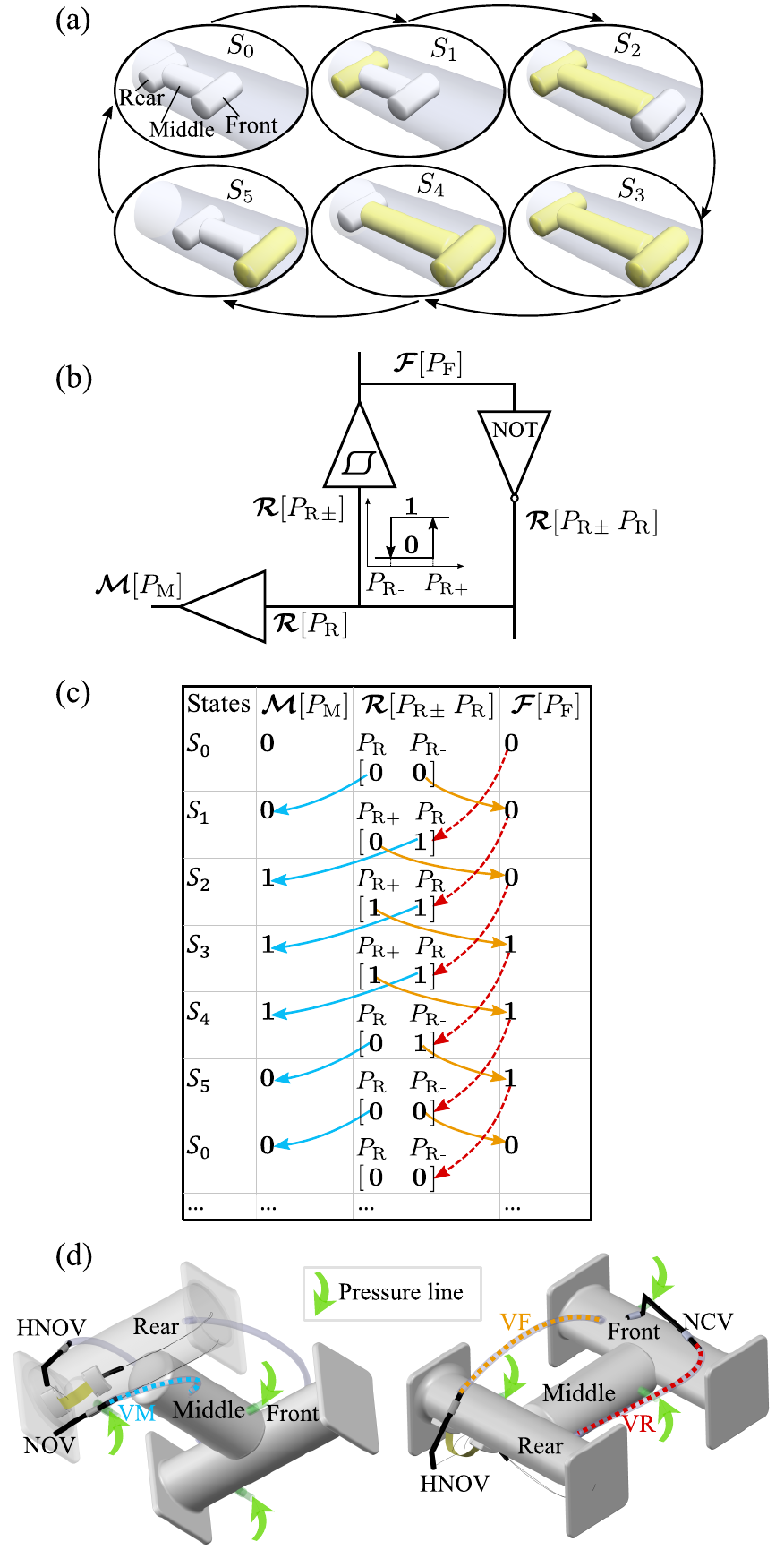}
\end{center}
\vspace{-1ex}
\caption{\textbf{Application: gait control for autonomous locomotion of a soft robot.} \textbf{(a)}~Illustration of the locomotion mechanism and gait requirements. The inflated actuators are shown in \textit{yellow}. \textbf{(b)}~Logic circuit devised to achieve the required actuator coordination for the gait in (a). \textbf{(c)}~State diagram use to predict the actuator response sequences corresponding to the logic circuit in (\ref{EQN07}) and diagram in~(b). \textbf{(d)}~Hardware implementation of the six-state machine described by (a), (b) and (c), employing the soft components in Figs.\,\ref{FIG02}~and~\ref{FIG03}. \label{FIG06}} 
\vspace{-2ex}
\end{figure}
\begin{figure*}[t!]
\vspace{1.6ex}
\begin{center}
\includegraphics[width=0.98\textwidth]{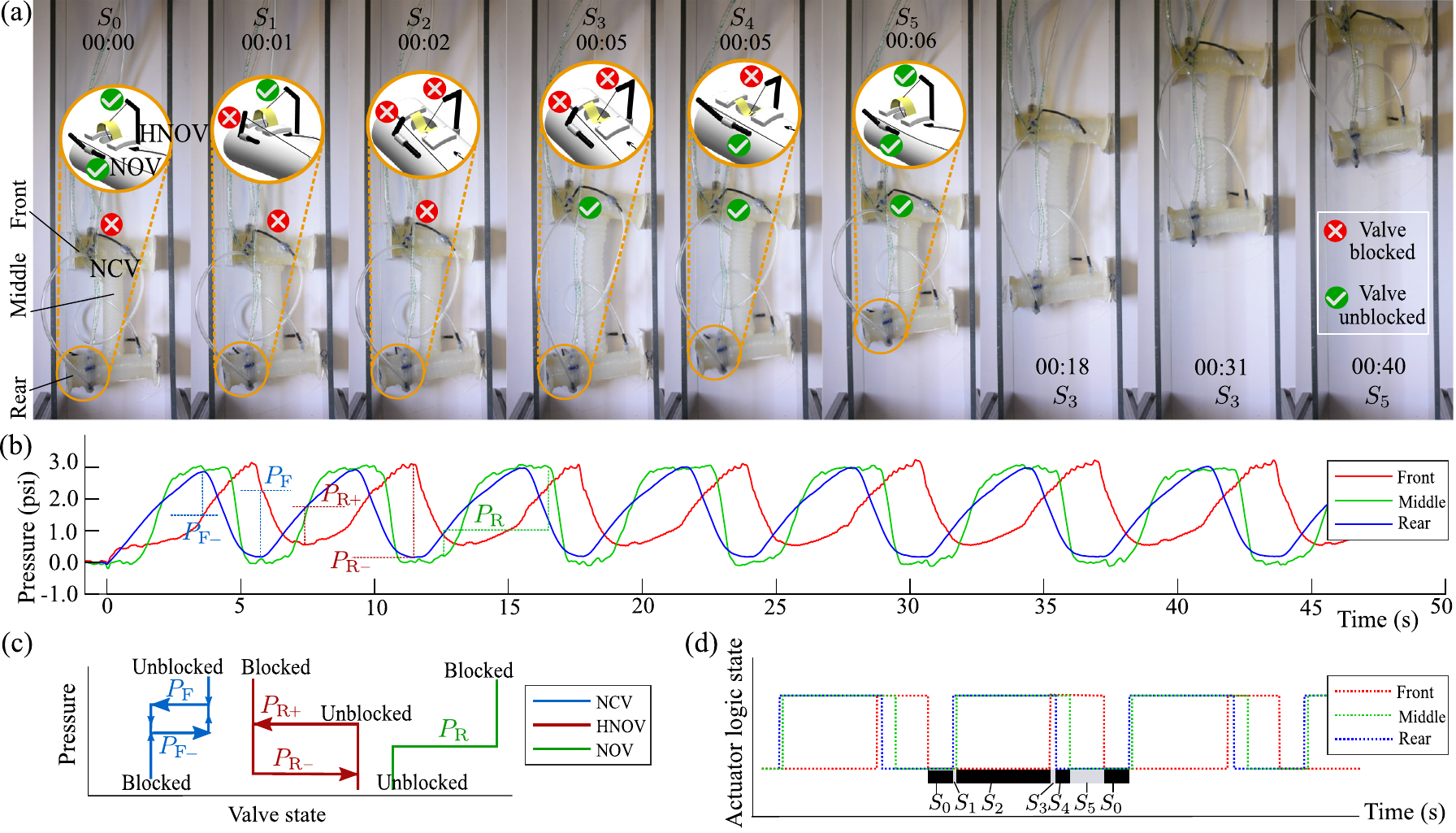}
\end{center}
\vspace{-1ex}
\caption{\textbf{Experimental results.} \textbf{(a)}~Photographic sequence that shows the crawler locomoting on a horizontal surface. Here, a complete gait cycle, from $S_0$ to $S_5$, is shown in the first six photo stills; illustration insets were added to visualize the binary states of the HNOV and NOV driven by the rear foot. \textbf{(b)}~Analog air pressure signals measured inside the three actuators of the robot. Note that these signals are used for identification and monitoring purposes only and not for control. \textbf{(c)}~Designed responses of the three valves of the robot. The threshold pressures of the valves were identified from the switching points of the experimental analog signals. The specific values are: $P_\ts{F-} = 1.5\,\ts{psi}$; $ P_\ts{F} = 2.3\,\ts{psi}$; $P_\ts{R-} = 0\,\ts{psi}$; $ P_\ts{R+} = 1.8\,\ts{psi}$; $P_\ts{R} = 1.1\,\ts{psi}$. \textbf{(d)}~Discretized logic state signals of the actuators. \label{FIG07}}  
 \vspace{-2ex}
\end{figure*}

When considering the control logic of the middle actuator, it is straightforward to see that a buffer logic that relates the middle actuator with the front foot, according to $\bs{\mathcal{M}}_{n+1}[P_\ts{M}] = \bs{\mathcal{R}}_n[P_\ts{R}]$, enables the middle actuator to output the desired logic state sequence in Table\,\ref{TABLE01}-(a). Then, after the logic structure for the entire robotic system is put together, we obtain the logic diagram in Fig.\,\ref{FIG06}-(b) and equations
\begin{align}
\begin{split}
\bs{\mathcal{R}}_{n+1}[P_{\ts{R}\pm}\: P_\ts{R}] &= \bs{\bar{\mathcal{F}}}_n[P_\ts{F}] \\
\bs{\mathcal{F}}_{n+1}[P_\ts{F}] &=\bs{\mathcal{R}}_n[P_{\ts{R}\pm}] \\
\bs{\mathcal{M}}_{n+1}[P_\ts{M}] &= \bs{\mathcal{R}}_n[P_\ts{R}] 
\end{split} \hspace{1.0ex},
\label{EQN07}
\end{align}
in which the logic state of the rear foot outputted from the NOT gate is ternary in terms of both the hysteretic threshold $P_{\ts{R}\pm}$ and the constant threshold $P_\ts{R}$; the two bits of this ternary state, in turn, function as two binary inputs for the two buffers. The overall logic structure and predicted logic response sequences of the actuators can be summarized using the state diagram in Fig.\,\ref{FIG06}-(c), which satisfies the actuator coordination requirements in Table\,\ref{TABLE01}-(a). Consistent with (\ref{EQN07}) and the diagram in Fig.\ref{FIG06}-(b), the dashed arrows denote the NOT logic and the solid arrows denote the two buffers. The inputs and outputs of the logic gates are indicated by the tails and heads of the arrows, respectively. Note that the logic operations in the state diagram of Fig.\,\ref{FIG06}-(c) follow the rules defined in Section\,\ref{SECTION03}.  

\vspace{-0.5ex}
\subsection{Implementation and Experimental Results}
\vspace{-0.5ex}
\label{SUBSECTION04C}
By cascading logic gates according to (\ref{EQN07}), the desired state machine can be physically realized. The resulting robotic design, composed of three mechanically-coupled soft oscillators, is depicted in Fig.\,\ref{FIG06}-(d). The first line in (\ref{EQN07}) represents a NOT gate in which an NC valve (NCV in Fig.\,\ref{FIG06}-(d)), with the threshold pressure $P_\ts{F}$, is driven by the front foot and connected to the vent line of the rear foot (VR in \textit{red} in Fig.\,\ref{FIG06}-(d)). To tune the threshold of the valve, the distance $s_0$ of the NCV is adjusted so that the VR is unblocked only when the pressure of the front foot exceeds $P_\ts{F}$ (i.e., the front foot is anchored to the trench). The second line in (\ref{EQN07}) represents a buffer in which a hysteretic NO valve (HNOV in Fig.\,\ref{FIG06}-(d)) is driven by the rear foot and connected to the vent line of the front actuator (VF in \textit{gold} in Fig.\,\ref{FIG06}-(d)). The threshold pressure of the valve takes either the value of $ P_{\ts{R}+} $ or $ P_{\ts{R}-} $, depending on whether the rear actuator is inflating or deflating. When tuning the higher threshold pressure of the HNOV, the string length is adjusted so that the bistable membrane snap-closes the VF when the pressure of the rear foot exceeds $P_{\ts{R}+}$, at which point the rear foot is reliably anchored. To tune the lower threshold, the distance between the two ends of the membrane are adjusted so that the bistable mechanism is forced to snap back only when the rear foot has sufficiently retracted to detach from the trench. The third line in (\ref{EQN07}) represents a buffer in which an NO valve (NOV in Fig.\,\ref{FIG06}-(d)), with a threshold pressure $P_\ts{R}$, is driven by the rear actuator and connected to the vent of the middle actuator (VM in \textit{blue} in Fig.\,\ref{FIG06}-(d)). To tune the NOV, the length of the string of the valve is adjusted so that the VM is blocked only when the pressure of the rear foot exceeds $P_\ts{R}$, the pressure value required to just anchor. Note that the decentralized control architecture of the system allows us to tune the responses of the feet first, while the pressure supply to the torso remains inactive; in the next step, the tuning of the torso response does not disrupt the already coordinated feet, as they do not require information from the torso to function.

To test the performance of the gait controller, we ran experiments in which the soft robot crawls inside a trench that was built with a horizontal smooth surface and a pair of poly-carbonate parallel plates (shown in Fig.\,\ref{FIG07}-(a)). In this case, three mini air pumps (Airpon\,D2028) supply the air pressures that feed the three actuators of the robot. In order to generate similar flow-rate inputs, the pumps that feed the feet share the same DC power supply while the pump that feeds the torso uses a separate DC power supply. MATLAB was used to analyze and visualize the pressure measurements collected using analog silicon sensors (Honeywell\,ASDX) connected to an Arduino processor. The achieved functionality and performance are demonstrated using the data in Fig.\,\ref{FIG07} and supplementary movie\,S1.mp4 \mbox{(\url{http://www.uscamsl.com/resources/RoboSoft2020/S1.mp4})}. During this test, the robot locomotes at an average speed of $0.383$\,m/min.

As seen in Fig.\,\ref{FIG07}-(a) and movie S1.mp4, whenever the system enters a state $S_n$ in Fig.\,\ref{FIG06}-(a), the air inputs and, hence, the dynamics of the robot are \textit{immediately} switched by the actions of the passive soft valves. As a result, the state of the robot immediately moves towards and then \textit{reaches} the desired next state $S_{n+1}$. The corresponding rhythmic analog pressure outputs from the three actuators are shown in Fig.\,\ref{FIG07}-(b). Note that the turning points of these signals indicate the switching actions of the valves and, hence, threshold pressures that are used to construct the corresponding valve states, which are plotted in Fig.\,\ref{FIG07}-(c). The actual response of the NCV driven by the front actuator switches at two threshold pressures, $P_{\ts{F}-} = 1.5$\,psi  and $P_\ts{F} = 2.3$\,psi. This observation indicates that, in this case, the valve dynamics has a set of intermediate states between a fully unblocked and reliably blocked status. However, this small switching error is tolerated by the system and actually contributes to a faster transition from $S_2$ (when the rear foot is anchored) to $S_4$ (when the front foot is anchored), i.e., a shorter duration of the state $S_3$. In this case, the identified threshold pressures $P_\ts{R}$, $P_\ts{F}$ and $P_\ts{M}$ are used to define the instantaneous logic states associated with the analog signals of the actuators, which are plotted in Fig.\,\ref{FIG07}-(d). \textit{Black} and \textit{gray} bars mark the system evolution in terms of the sequence of finite states from $S_0$ to $S_5$. These results indicate consistency with the state machine requirements in Table\,\ref{TABLE01}-(a).

\vspace{-0.5ex}
\section{Conclusions}
\vspace{-0.5ex}
\label{SECTION05}
We presented a new approach to the design, modeling and implementation of integrated electronics-free localized logic controllers for pneumatic soft robots. To enable this controller synthesis approach, we developed new types of soft switch-valves that can be embedded distributedly in the skins of a soft robot and thus be used to couple the states of the composing actuators, employing simple fine-tuning features. In the proposed method, binary (or multi-level) logic states for the soft actuators of a robot are defined by processing the corresponding analog actuator-pressure signals using threshold functions. These functions are defined in terms of one or multiple threshold values, with or without hysteresis, depending on the design requirements for the robot; by design, the actuators do not need to actually generate discrete responses as, for example, in\,\cite{8722820}. The integration of soft components define logic gates characterized by the dynamical interactions introduced between the actuators by the switch-valves; consistently, desired coordination patterns for the analog actuator states are formulated as finite-state machines. In this way, principles for digital design can be applied to derive the distributed mechanical logic controllers required to coordinate the actuators of a soft robot in order to follow a desired oscillation pattern of operation. As an example, we presented a soft robotic design that requires simple fabrication and tuning (Section\,\ref{SECTION04}), in which a multi-valued logic enabled the creation of a decentralized control structure using a small number of interconnected switch-valves. The resulting system emulates the decentralized control mechanisms observed in animal muscles. We anticipate that this proposed approach will enable the development of soft robots composed of a great number of modules.

\bibliographystyle{IEEEtran}
\balance
\bibliography{mybib}

\begin{thebibliography}{10}
\providecommand{\url}[1]{#1}
\csname url@samestyle\endcsname
\providecommand{\newblock}{\relax}
\providecommand{\bibinfo}[2]{#2}
\providecommand{\BIBentrySTDinterwordspacing}{\spaceskip=0pt\relax}
\providecommand{\BIBentryALTinterwordstretchfactor}{4}
\providecommand{\BIBentryALTinterwordspacing}{\spaceskip=\fontdimen2\font plus
\BIBentryALTinterwordstretchfactor\fontdimen3\font minus
  \fontdimen4\font\relax}
\providecommand{\BIBforeignlanguage}[2]{{%
\expandafter\ifx\csname l@#1\endcsname\relax
\typeout{** WARNING: IEEEtran.bst: No hyphenation pattern has been}%
\typeout{** loaded for the language `#1'. Using the pattern for}%
\typeout{** the default language instead.}%
\else
\language=\csname l@#1\endcsname
\fi
#2}}
\providecommand{\BIBdecl}{\relax}
\BIBdecl

\bibitem{shepherd2013using}
{R. F. Shepherd}, {A. A. Stokes}, {J. Freake}, {J. Barber}, {P. W. Snyder}, {A.
  D. Mazzeo}, {L. Cademartiri}, {S. A. Morin}, and {G. M. Whitesides}, ``{Using
  Explosions to Power a Soft Robot},'' \emph{{Angew. Chem.}}, vol.~52, no.~10,
  pp. 2892--2896, Mar. 2013.

\bibitem{unger2000monolithic}
{M. A. Unger}, {H.-P. Chou}, {T. Thorsen}, {A. Scherer}, and {S. R. Quake},
  ``{Monolithic Microfabricated Valves and Pumps by Multilayer Soft
  Lithography},'' \emph{Science}, vol. 288, no. 5463, pp. 113--116, Apr. 2000.

\bibitem{mosadegh2010integrated}
{B. Mosadegh}, {C.-H. Kuo}, {Y.-C. Tung}, {Y.-s. Torisawa}, {T. Bersano-Begey},
  {H. Tavana}, and {S. Takayama}, ``{Integrated Elastomeric Components for
  Autonomous Regulation of Sequential and Oscillatory Flow Switching in
  Microfluidic Devices},'' \emph{Nat. Phys.}, vol.~6, no.~6, pp. 433--437, Jun.
  2010.

\bibitem{rothemund2018soft}
{P. Rothemund}, {A. Ainla}, {L. Belding}, {D. J. Preston}, {S. Kurihara}, {Z.
  Suo}, and {G. M. Whitesides}, ``{A Soft, Bistable Valve for Autonomous
  Control of Soft Actuators},'' \emph{Sci. Robot.}, vol.~3, no.~16, p.
  eaar7986, Mar. 2018.

\bibitem{ahrar2015programmable}
{S. Ahrar}, ``{A Programmable Microfluidic Finite State Machine for the
  Autonomous Lab on a Chip},'' Ph.D. dissertation, UC Irvine, 2015.

\bibitem{wehner2016integrated}
{M. Wehner}, {R. L. Truby}, {D. J. Fitzgerald}, {B. Mosadegh}, {G. M.
  Whitesides}, {J. A. Lewis}, and {R. J. Wood}, ``{An Integrated Design and
  Fabrication Strategy for Entirely Soft, Autonomous Robots},'' \emph{Nature},
  vol. 536, no. 7617, pp. 451--455, Aug. 2016.

\bibitem{mahon2019soft}
{S. T. Mahon}, {A. Buchoux}, {M. E. Sayed}, {L. Teng}, and {A. A. Stokes},
  ``{Soft Robots for Extreme Environments: Removing Electronic Control},'' in
  \emph{{Proc. 2019 IEEE Int. Conf. Soft Robot. (RoboSoft\,2019)}}, Seoul,
  Korea, Apr. 2019, pp. 782--787.

\bibitem{preston2019digital}
{D. J. Preston}, {P. Rothemund}, {H. J. Jiang}, {M. P. Nemitz}, {J. Rawson},
  {Z. Suo}, and {G. M. Whitesides}, ``{Digital Logic for Soft Devices},''
  \emph{PNAS}, vol. 116, no.~16, pp. 7750--7759, Apr. 2019.

\bibitem{zhang2017logic}
{Q. Zhang}, {M. Zhang}, {L. Djeghlaf}, {J. Bataille}, {J. Gamby}, {A.-M.
  Haghiri-Gosnet}, and {A. Pallandre}, ``{Logic Digital Fluidic in Miniaturized
  Functional Devices: Perspective to the Next Generation of Microfluidic
  Lab-on-Chips},'' \emph{Electrophoresis}, vol.~38, no.~7, pp. 953--976, Apr.
  2017.

\bibitem{rhee2009microfluidic}
{M. Rhee} and {M. A. Burns}, ``{Microfluidic Pneumatic Logic Circuits and
  Digital Pneumatic Microprocessors for Integrated Microfluidic Systems},''
  \emph{{Lab Chip}}, vol.~9, no.~21, pp. 3131--3143, Nov. 2009.

\bibitem{devaraju2012pressure}
{N. S. G. K. Devaraju} and {M. A. Unger}, ``{Pressure Driven Digital Logic in
  PDMS Based Microfluidic Devices Fabricated by Multilayer Soft Lithography},''
  \emph{Lab Chip}, vol.~12, no.~22, pp. 4809--4815, Nov. 2012.

\bibitem{weaver2010static}
{J. A. Weaver}, {J. Melin}, {D. Stark}, {S. R. Quake}, and {M. A. Horowitz},
  ``{Static Control Logic for Microfluidic Devices Using Pressure-Gain
  Valves},'' \emph{{Nat. Phys.}}, vol.~6, no.~3, pp. 218--223, Jan. 2010.

\bibitem{kai_luo_2019}
{K. Luo}, {P. Rothemund}, {G. M. Whitesides}, and {Z. Suo}, ``{Soft Kink
  Valves},'' \emph{J. Mech. Phys. Solids}, vol. 131, pp. 230--239, Oct. 2019.

\bibitem{ariel_robio_2016}
{A. A. Calder\'on}, {J. C. Ugalde}, {J. C. Zagal}, and {N. O.
  P\'erez-Arancibia}, ``{Design, Fabrication and Control of a
  Multi-Material--Multi-Actuator Soft Robot Inspired by Burrowing Worms},'' in
  \emph{{Proc. 2016 IEEE Int.\,Conf.\,Robot.\,Biomim.\,(RoBio\,2016)}},
  Qingdao, China, Dec. 2016, pp. 3--7.

\bibitem{joey_robio_2017}
{J. Z. Ge}, {A. A. Calder\'on}, and {N. O. P\'erez-Arancibia}, ``{An
  Earthworm-Inspired Soft Crawling Robot Controlled by Friction},'' in
  \emph{{Proc. 2017 IEEE Int.\,Conf.\,Robot.\,Biomim.\,(RoBio\,2017)}}, Macau
  SAR, China, Dec. 2017, pp. 834--841.

\bibitem{joey2019earthworm}
{J. Z. Ge}, {A. A. Calder{\'o}n}, {L. Chang}, and {N. O. P{\'e}rez-Arancibia},
  ``{An Earthworm-Inspired Friction-Controlled Soft Robot Capable of
  Bidirectional Locomotion},'' \emph{Bioinspir. Biomim.}, vol.~14, no.~3, p.
  036004, May 2019.

\bibitem{calderon2019earthworm}
{A. A. Calderon}, {J. C. Ugalde}, {L. Chang}, {J. C. Zagal}, and {N. O.
  P\'erez-Arancibia}, ``{An Earthworm-Inspired Soft Robot with Perceptive
  Artificial Skin},'' \emph{Bioinspir. Biomim.}, vol.~14, no.~5, p. 056012,
  Sep. 2019.

\bibitem{8722820}
{A. J. Partridge} and {A. T. Conn}, ``{Buckling Elements for Elastomer
  Deformation},'' in \emph{{Proc. 2019 IEEE Int. Conf. Soft Robot.
  (RoboSoft\,2019)}}, Seoul, Korea, Apr. 2019, pp. 68--73.

\end{thebibliography}

\end{document}